%% file: phase2.tex
\documentclass[lettersize,journal]{IEEEtran}
\usepackage{amsmath,amsfonts}
\usepackage{algorithmic}
\usepackage{array}
\usepackage[caption=false,font=normalsize,labelfont=sf,textfont=sf]{subfig}
\usepackage{textcomp}
\usepackage{stfloats}
\usepackage{url}
\usepackage{verbatim}
\usepackage{graphicx}
\def\BibTeX{{\rm B\kern-.05em{\sc i\kern-.025em b}\kern-.08em
    T\kern-.1667em\lower.7ex\hbox{E}\kern-.125emX}}
\usepackage{balance}

\def\eg{\emph{e.g. }}

\def\etc{\emph{etc}}

\usepackage{amsmath}
\usepackage{booktabs}
\usepackage{amssymb}
\usepackage{bbding}
\usepackage{mathtools}
\usepackage{amsthm}
\usepackage{gensymb}
\usepackage{pifont}
\def\CircleArrowright{\ensuremath{
  \rotatebox[origin=c]{310}{$\circlearrowright$}}}
\newcommand{\rvlnbert}{VLN$\protect\CircleArrowright$BERT}

\usepackage{multirow}
\usepackage{graphicx}
\usepackage{xcolor}

\usepackage{hyperref}
\usepackage{cleveref}
\usepackage{algorithmic}
\usepackage{algorithm}


\begin{document}
\title{FlexVLN: Flexible Adaptation for Diverse Vision-and-Language Navigation Tasks}
\author{Siqi Zhang, Yanyuan Qiao, Qunbo Wang, Longteng Guo, Zhihua Wei, Jing Liu,~\IEEEmembership{Member,~IEEE}
\thanks{This work was done when Siqi Zhang was an intern at the Institute of Automation, Chinese Academy of Sciences.
This work was supported in part by the Artificial Intelligence-National Science and Technology Major Project (NO. 2023ZD0121200), in part by the National Nature Science Foundation of China: 62376199, 62206170, 62206279, and 62437001.
\textit{(Corresponding author: Zhihua Wei and Jing Liu.)}}
\thanks{S. Zhang and Z. Wei are with the Department of Computer Science and Technology, Tongji University, Shanghai 201804, China. (e-mail: 2211172@tongji.edu.cn, zhihua\_wei@tongji.edu.cn)}
\thanks{Y. Qiao is with the Australian Institute for Machine Learning, School of Computer Science, The University of Adelaide, Adelaide, SA 5005, Australia. (e-mail: yanyuan.qiao@adelaide.edu.au)}
\thanks{Q. Wang, L. Guo, J. Liu are with the Institute of Automation, Chinese Academy of Sciences, Beijing 100190, China, while J. Liu is also with the School of Artificial Intelligence, University of Chinese Academy of Science, Beijing 100190, China. (e-mail: qunbo.wang@ia.ac.cn, longteng.guo@nlpr.ia.ac.cn, jliu@nlpr.ia.ac.cn)}
}

\maketitle

\begin{abstract}
The aspiration of the Vision-and-Language Navigation (VLN) task has long been to develop an embodied agent with robust adaptability, capable of seamlessly transferring its navigation capabilities across various tasks.
Despite remarkable advancements in recent years, most methods necessitate dataset-specific training, thereby lacking the capability to generalize across diverse datasets encompassing distinct types of instructions.
Large language models (LLMs) have demonstrated exceptional reasoning and generalization abilities, exhibiting immense potential in robot action planning.
In this paper, we propose FlexVLN, an innovative hierarchical approach to VLN that integrates the fundamental navigation ability of a supervised-learning-based Instruction Follower with the robust generalization ability of the LLM Planner, enabling effective generalization across diverse VLN datasets. 
{Moreover, a verification mechanism and a multi-model integration mechanism are proposed to mitigate potential hallucinations by the LLM Planner and enhance execution accuracy of the Instruction Follower.}
We take REVERIE, SOON, {and CVDN-target} as out-of-domain datasets for assessing generalization ability. The generalization performance of FlexVLN surpasses that of all the previous methods to a large extent.
\end{abstract}


\begin{figure*}
    \centering
    \includegraphics[width=0.98\linewidth]{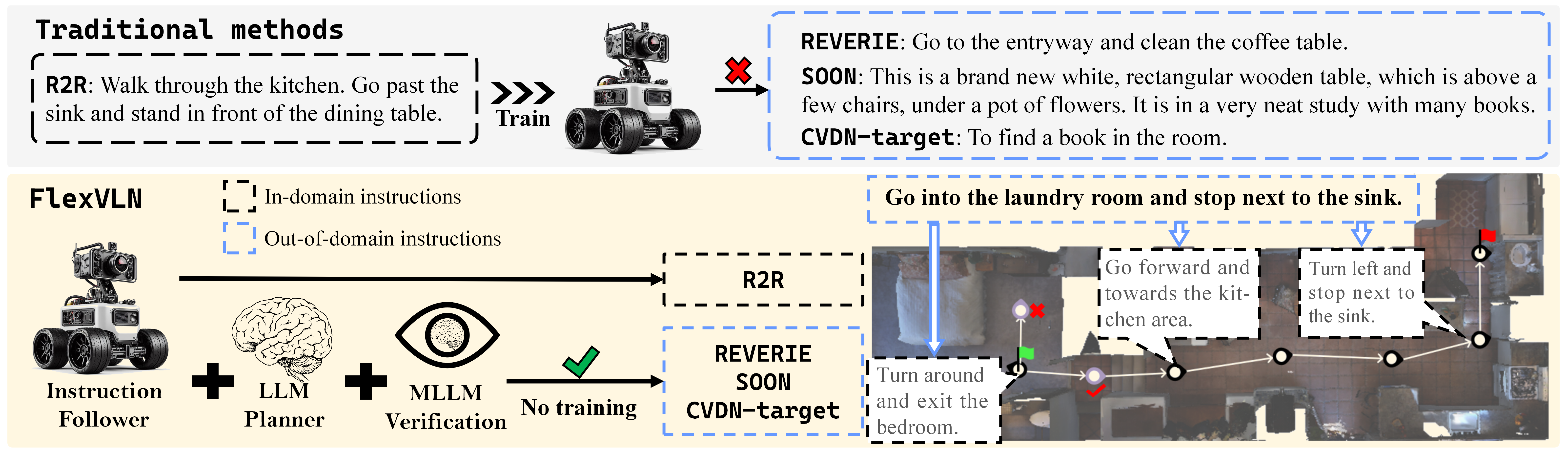}
    \vspace{-7pt}
    \caption{{Comparison between traditional VLN methods and FlexVLN. Traditional methods are usually trained on specific datasets and suffer low success rates when generalizing to other datasets. 
    In contrast, our FlexVLN demonstrates exceptional generalization performance without necessitating additional training or fine-tuning efforts on the target dataset. 
    An illustrative example is presented in the bottom right, where starting from the green flag, the LLM Planner transforms the out-of-domain instruction into step-by-step guidances (validated for feasibility by the MLLM) for seamless execution by the Instruction Follower, ultimately leading to the destination (red flag). 
    \includegraphics[width=1.3ex,height=1.5ex]{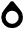} denotes the navigation node and heading.
    The Instruction Follower employs an ensemble of three models. When their action decisions diverge (as indicated by the purple symbols), a LLM is employed to determine the optimal action based on guidance.}
    }
    \label{fig:teaser}
    \vspace{-13pt}
\end{figure*}

\section{Introduction}
Vision-and-Language Navigation (VLN) ~\cite{anderson2018vision,gu22vlnsurvey} requires an agent to understand natural language instructions, perceive the visual world, and perform navigation actions to reach a target location.
It has significant potential in real-world applications such as domestic assistant robots and has garnered widespread interest. Consequently, numerous benchmark datasets have been proposed, each emphasizing distinct categories of instructions. For instance, R2R~\cite{anderson2018vision} concentrates on evaluating an agent's ability in following step-by-step instructions;
RxR~\cite{ku2020room} emphasizes longer and multilingual instructions; CVDN~\cite{thomason2020vision} focuses on dialogue-based instructions; 
REVERIE~\cite{qi2020reverie} and SOON~\cite{zhu2021soon} assess the agent's capability to explore the environment and locate remote objects using high-level instructions.

%
%

Recent studies in VLN have witnessed a remarkable surge in the development of supervised learning methods~\cite{hao2020towards,hop,chen2021history,an2023bevbert,wang2023gridmm}, leading to significant advancements in performance across diverse datasets.
However, despite incorporating pretraining strategies, these methods still necessitate separate fine-tuning for each dataset. As a result, trained agents are confined to a specific dataset and face challenges when it comes to generalizing across other datasets.
For instance, ScaleVLN~\cite{wang2023scaling} trained on R2R dataset exhibits strong instruction-following ability and achieves close-to-human performance. Nevertheless, it still requires fine-tuning or even re-pretraining on the target dataset in order to effectively handle high-level instructions in REVERIE and SOON.
In essence, even state-of-the-art supervised approaches lack the capability to generalize across various VLN datasets.
%
The generalization ability of agents has always been a crucial research direction, as it plays a pivotal role in facilitating the agent's future practical deployment, where the agent is expected to undergo training once and then be applied to diverse VLN tasks. Most of the past works focused on the generalization research on unseen environments~\cite{panogen,lily}, while generalizing different types of instructions has been overlooked.

%
%
Benefiting from the unprecedented scale of the training corpus, large language models (LLMs) exhibit robust reasoning and generalization abilities, leading to the emergence of LLM-based VLN methods.
The pioneering NavGPT~\cite{zhou2023navgpt} tried to leverage the reasoning ability of GPT to develop a purely LLM-based instruction following agent.
On this basis, DiscussGPT~\cite{long2023discuss} proposes engaging in discussions with eight experts to collect essential information before each movement.
MapGPT~\cite{chen2024mapgpt} converts the topological map constructed online into prompts to encourage global exploration.
Although these methods are not confined to a specific VLN dataset, their performance is significantly inferior to supervised learning methods. Moreover, due to the dynamic nature of the VLN task, LLM needs to be applied at every single step of navigation, resulting in substantial time and resource consumption.
Furthermore, their utilization of LLMs is suboptimal. 
Firstly, these methods require the LLM to predict the next action directly by selecting a node from current neighboring nodes. However, the LLM is only aware of each node's orientation and lacks knowledge about their specific locations. Thus, it cannot differentiate between two distinct nodes in the same orientation.
Secondly, purely LLM-based agents have difficulties when it comes to navigating around obstacles without deviating from their intended path. For instance, suppose an agent is attempting to exit through a door in its front view but encounters obstruction from a kitchen island that prevents any viable passage in the desired direction. This situation may result in deviation from the intended trajectory as the agent seeks alternative routes.


%
%

In this paper, we introduce FlexVLN, an innovative hierarchical navigation policy that integrates fundamental navigation capability with the capacity for generalization across a variety of VLN tasks.
FlexVLN consists of {three main components: the LLM Planner for high-level planning and generating fine-grained guidances; the Instruction Follower for executing guidances to retain basic navigation capability; and the Object Locator for locating the target object at the end of the navigation process.
To address potential hallucinations, a verification mechanism is proposed to validate the feasibility of the guidance generated by the LLM Planner through a MLLM. 
Moreover, in order to enhance the execution accuracy of the Instruction Follower, we propose a multi-model integration mechanism, where three models are ensembled. At each time step, if there are inconsistencies in action decisions among these models, a LLM will be consulted for assistance in determining optimal actions based on the guidance.}
This strategy enables FlexVLN to adapt to VLN datasets outside the domain effectively.
As depicted in \Cref{fig:teaser}, traditional methods trained on a specific dataset fail to generalize to other datasets. Comparatively, our FlexVLN demonstrates superior generalization performance without requiring additional training data or fine-tuning efforts.
To the best of our knowledge, we are the first to address the challenge of generalization across different VLN tasks without any additional training.
%
%

More specifically, as our {Instruction Follower incorporates models that are trained on fine-grained instructions,} we consider instructions that deviate from this style as out-of-domain (OOD) instructions.
As exemplified in the bottom right of \Cref{fig:teaser},
given an OOD instruction, the LLM Planner conducts high-level planning by inferring where to go next based on current observations and navigation history. It then transforms this plan into a fine-grained guidance that aligns with the familiarity of the Instruction Follower.
{
Subsequently, the feasibility of the guidance is validated by a MLLM. In case of infeasibility, feedback on the reasons will be provided to the LLM Planner for re-planning and re-generating a new guidance. If feasible, the Instruction Follower executes the guidance, which may involve multiple low-level navigational actions for completion.
Three models are ensembled in the Instruction Follower. At each time step, if their actions are consistent, the action is executed; otherwise, assistance is sought from a LLM. The LLM selects the optimal action based on the guidance.
}
Upon completion, the Instruction Follower seeks further guidance from the LLM Planner. This process iterates until the LLM Planner confirms that the destination has been reached.
%
By alternating high-level planning and low-level execution, the frequency of LLM calls can be significantly reduced while optimizing the utilization of LLM.
This is achieved by assigning the LLM to specify precise destinations instead of requiring them to select nodes with imprecise locations, and executing plans using the Instruction Follower to ensure they remain unchanged while navigating around obstacles.
As a result, the challenges posed by indistinguishability between nodes in the same orientation and path deviation during navigation around obstacles are effectively overcome.

{
In this paper, we focus on the generalizability on high-level instructions, which align more closely with real-world applications. We conduct experiments on three distinct VLN benchmarks REVERIE~\cite{qi2020reverie}, SOON~\cite{zhu2021soon}, and CVDN-target~\cite{qiao2025copilot} to validate the effectiveness of our proposed method.
}
The generalization performance of FlexVLN surpasses that of all previous methods on the target dataset, achieving comparable results to models specifically trained on the target dataset.
In summary, our contributions are as follows:
\begin{itemize}
    \item {We introduce FlexVLN, an innovative hierarchical navigation system that tackles the challenge of generalization across diverse VLN tasks without any additional training.}
    \item {We propose a multi-step pipeline to seamlessly integrate the generalization ability of the LLM and the foundational navigation ability of the Instruction Follower. A verification mechanism and a multi-model integration mechanism are proposed to mitigate potential hallucinations and enhance guidance execution accuracy.}
    \item {FlexVLN demonstrates impressive generalization performance on REVERIE, SOON, and CVDN-target, significantly surpassing previous methods.}
\end{itemize}

\section{Related Work}
\subsection{Vision-and-Language Navigation (VLN)}
Embodied navigation can be broadly classified into two categories: visual navigation (VN)~\cite{borenstein1989real,borenstein1991vector} and language-driven visual navigation (VLN)~\cite{krantz2020beyond,zhu2022diagnosing,liang2023mo,li2021improving,hwang2023meta,gao2023adaptive,wu2024vision,zhang2024vision}. VN primarily focuses on mapping and localization~\cite{dissanayake2001solution,duh2020v,piao2019real}, while VLN places more emphasis on natural-language comprehension and environment exploration.
VLN is essential for versatile embodied navigation agents and has received increasing attention from multimodal understanding and robotic communities.
A large amount of VLN benchmark datasets have been proposed, featuring diverse types of instructions and assessing various capabilities of the navigation agent. The first VLN benchmark Room-to-room (R2R)~\cite{anderson2018vision} provides step-by-step instructions, necessitating precise execution of commanded actions in sequential order. Room-for-room (R4R)~\cite{jain2019stay} evaluates the agent's fidelity by increasing instruction length. Room-across-room (RxR)~\cite{ku2020room} incorporates instructions in three languages to assess the agent's proficiency in comprehending varied linguistic contexts. CVDN~\cite{thomason2020vision} instructs the agent in the form of dialogues. Both REVERIE~\cite{qi2020reverie} and SOON~\cite{zhu2021soon} are object-oriented benchmarks, requiring exploration of the environment, navigation towards designated locations, and location of remote objects. 
{R2R-CE~\cite{krantz2020beyond} and RxR-CE migrate R2R and RxR to continuous scenarios, necessitating agents to perform low-level actions.}

%
A substantial body of prior research mainly tackles the VLN task in the following ways: {
topological graph-based methods~\cite{hong2020language,deng2020evolving,chen2021topological,chen2022think}, 
methods based on top-down semantic map~\cite{georgakis2022cross,irshad2022semantically,chen2022weakly,hong2023learning},
pre-training and transformer-based methods~\cite{an2023bevbert,hong2020rvlnbert,guhur2021airbert,lovis2022,cui2023grounded,qiao2022hop,hop}, 
reinforcement learning techniques~\cite{wang2019reinforced}, 
environment augmentation strategies~\cite{chen2022learning,li2022envedit,panogen}, 
curriculum learning approaches~\cite{babywalk,hong2020sub,he2021landmark}, 
and the utilization of external knowledge~\cite{li2023kerm}.}
%
However, these approaches are dataset specific~\cite{dorbala2022clip}. That is to say, the generalization of a model trained on one dataset to another is challenging and may result in significant performance degradation, even if both datasets are built on the same simulator. These conventional training methods fall short of meeting the requirements for practical applications where instructions can vary extensively.
Despite extensive efforts in instruction augmentation methods~\cite{fu2020counterfactual,dou2022foam,wang2022less,wang2022counterfactual,zeng2023kefa,zhang2023vln}, the synthetic instructions remain constrained by the style of training samples and fail to capture the diverse range of language instructions required for realistic applications.
In this work, we aim to develop an agent that can flexibly generalize across diverse VLN datasets through a hierarchical navigation system. Our FlexVLN combines the generalization ability of the LLM Planner with the fundamental navigation ability of the Instruction Follower, ensuring superior performance in handling out-of-domain instructions.

\vspace{-5pt}
\subsection{LLMs in Embodied Navigation}
Building upon the strong generalization and reasoning ability of large pre-trained models~\cite{kojima2022large,zheng2022jarvis,10004507,10598361}, a handful of contemporary studies have begun to explore the utilization of generative models for embodied navigation in a modularized form.
LM-Nav~\cite{shah2023lm} utilizes GPT-3~\cite{brown2020language} for extracting landmarks, CLIP \cite{radford2021learning} for grounding, and a temporal distance predictor for navigation.
{Mic~\cite{qiao2023mic} leverages the commonsense knowledge in GPT-2~\cite{radford2019gpt2} to augment coarse-grained instructions with details regarding destination and ongoing action.
LangNav~\cite{pan2023langnav} converts VLN into a pure-text task, thereby tackling the challenge of data scarcity in the embodied domain.
VLN-Copilot~\cite{qiao2025copilot} enables the agent to proactively seek assistance from LLMs when encountering confusion.}
In addition to serving as a helper in LM-Nav, LLMs can also be used as a planner~\cite{ahn2022can,huang2022language}.
$A^2$Nav \cite{chen20232} utilizes GPT-3 for decomposing instructions into sub-tasks and executes them with multiple ZSON~\cite{majumdar2022zson} models.
ESC~\cite{zhou2023esc} applies Deverta-v3~\cite{he2021debertav3} to perform commonsense reasoning about the target object.
Furthermore, LLMs can directly execute actions.
NavGPT~\cite{zhou2023navgpt} generates actions directly with GPT-4~\cite{achiam2023gpt}. 
DiscussNav~\cite{long2023discuss} engages in discussions with multiple domain experts on this foundation.
MapGPT~\cite{chen2024mapgpt} converts the topological map constructed online into prompts to encourage global exploration.
LLM-planner~\cite{song2023llm} utilizes GPT-3 for generating high-level action that belongs to the given action space.
{NavCoT~\cite{lin2024navcot} prompts the LLM to forecast the navigational chain-of-thought, encompassing future observation imagination, probable observation selection, and action determination.}

The aforementioned methods, however, are all developed within the context of object navigation or instruction following.
To be specific, 
object navigation \cite{xia2018gibson,chang2017matterport3d,ramakrishnan2021habitat} involves exploring the environment to find a target object (\eg ``plant'') without any constraints. The instruction following \cite{anderson2018vision,ku2020room,jain2019stay} involves adhering to a step-by-step instruction (\eg ``Keep walking forward passing all the picture frames on the wall. Turn left at the corner before where the chairs are placed.''). 
Nevertheless, in practical applications, coarse-grained instructions (\eg ``Go to the lounge on first floor and turn on the lamp.'') are predominantly employed, which is more demanding than the previous two tasks.
In our method, we act the LLM as a high-level planner which is required to reason about the given instruction, navigation history, and current observations, and generate fine-grained guidance to assist the Instruction Follower in accomplishing diverse tasks.

\begin{figure*}[t]
    \centering
    \includegraphics[width=0.98\linewidth]{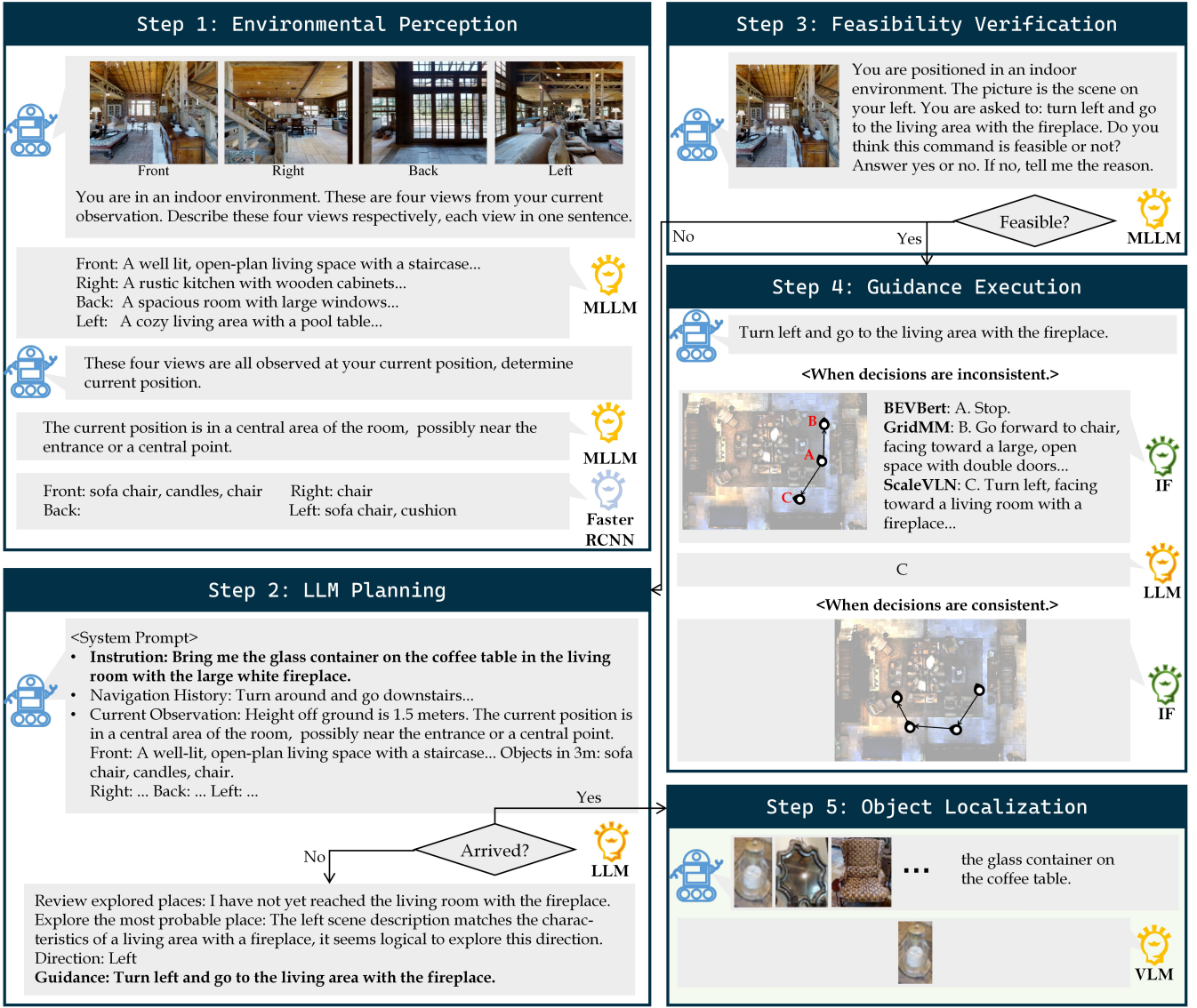}
    \vspace{-10pt}
    \caption{{FlexVLN facilitates the seamless integration of the LLM Planner and the Instruction Follower (IF) through a five-step pipeline. The LLM Planner utilizes the given OOD instruction, navigation history, and environmental observations obtained in \texttt{Step1} to determine an exploration strategy, generating a fine-grained guidance that aligns with the Instruction Follower's familiarity in \texttt{Step2}. The feasibility of the guidance is then verified in \texttt{Step3}. The Instruction Follower is responsible for executing low-level navigation actions based on this guidance in \texttt{Step4}. At the end of the navigation process, the Object Locator is tasked to locate the target object in \texttt{Step5}.}}
    \label{fig:framework}
    \vspace{-15pt}
\end{figure*}

\section{Problem Formulation}
\label{Problem formulation}
In the standard VLN setup for discrete environments \cite{anderson2018vision}, the environment is an undirected graph $\mathcal{G} = \{\mathcal{V}, \mathcal{E}\}$, where $\mathcal{V} = \{V_i\}^K_{i=1}$ denotes $K$ navigable nodes, and $\mathcal{E}$ denotes connectivity edges.
Given a natural language instruction $\mathcal{I}$, an agent is initialized at a starting node, and its goal is to interpret the instruction and to traverse the graph to the target location and find the object specified by the instruction.
At each time step $t$, the agent receives an observation $\mathcal{O}_t$, position coordinates ($x_t, y_t, z_t$), and neighboring nodes $\mathcal{N}(V_t)$ of its current node $V_t$, as well as agent's current heading $\theta_t$ and elevation angle $\phi_t$.
The observation comprises $N$ views $\mathcal{O}_t = \{o_i^t\}^N_{i=1}$, representing the egocentric perspectives of agents in varying orientations.
Throughout the navigation process, the agent's action space $\mathcal{A}_t$ includes moving to $V_{t+1} \in \mathcal N (V_t)$ and stopping at $V_t$.
The agent is required to predict an action $a_t \in \mathcal{A}_t$ based on the instruction $\mathcal{I}$, current observations $\mathcal{O}_t$, and navigation history $h_t$.
Once the agent decides to stop, it needs to locate the target object at the current node.

\section{Method}
Navigation systems necessitate two essential capabilities: strategic route planning based on instructions and execution of designated plans. Therefore, navigation inherently follows a hierarchical process. The supervised training methodology equips the model with both proficiencies through in-domain training. The LLM-based approaches facilitate simultaneous planning and execution of LLMs. However, it is worth noting that LLMs excel at planning while domain-trained models outperform in executing navigation actions.
Hence, we propose FlexVLN, a hierarchical navigation system that leverages the generalization ability, reasoning prowess, and commonsense knowledge of the LLM Planner to provide guidance for exploration while utilizing the low-level action execution capability of the Instruction Follower.

\subsection{Overview}
In this study, we focus on enhancing the agent's generalization ability across OOD VLN datasets. 
FlexVLN comprises three main components: the LLM Planner, which conducts high-level planning and transforms OOD instructions into fine-grained guidances that are familiar to the Instruction Follower; the Instruction Follower, which executes low-level navigation actions based on the provided guidances; and the Object Locator, which identifies the target object at the end of the navigation process.
{Moreover, a verification mechanism is proposed to mitigate potential hallucinations and unreasonable planning by the LLM Planner. For the Instruction Follower, a multi-model integration mechanism is proposed to enhance the accuracy of guidance execution.}

{As depicted in \Cref{fig:framework}, the pipeline of FlexVLN consists of five steps. 
\texttt{Step1} is environmental perception (\Cref{sec:env_percep}), where off-the-shelf models are applied to translate observations into scene and object descriptions.
Then the LLM Planner is applied in the second step (\Cref{sec:llm_plan}). The LLM Planner should engage in complex reasoning to generate a fine-grained guidance according to current observations and navigation history.
In order to mitigate hallucinations that occurred in \texttt{Step1} and the generation of unfeasible guidances by the LLM Planner, such as going through a closed door, we employ a MLLM in \texttt{Step 3} to verify the feasibility of the guidance (\Cref{sec:feasible}). If deemed unfeasible, return to \texttt{Step2} and provide feedback on infeasibility to the LLM Planner, prompting it to regenerate a guidance. If deemed feasible, proceed to \texttt{Step4}.
In \texttt{Step4} (\Cref{sec:execute}), three models are integrated as the Instruction Follower. During the implementation of guidance, if the actions determined by these three models are identical, they are considered accurate and executed. In case of discrepancies in their actions, a LLM is sought for help in judgment. The LLM is asked to select the optimal action according to the guidance and the Instruction Follower will execute this action.}
Upon completion, the trajectory of the Instruction Follower is converted into a textual format, fed back to the LLM Planner, and appended to the navigation history.
{This iterative process (\texttt{Step1} to \texttt{Step4}) continues until the LLM Planner determines that the destination has been reached.}
Finally, in \texttt{Step5}, the Object Locator is applied to pick out the target object (\Cref{sec:localize}).
Notably, the LLM Planner in FlexVLN is not applied at every time step, but rather exclusively employed upon completion of the previous guidance by the Instruction Follower. 
%

{The key distinction between our FlexVLN and previous LLM-based VLN methods lies in the fact that FlexVLN is a navigation system designed to effectively generalize across diverse datasets without requiring additional training, as opposed to employing LLM as a navigation agent for directly taking low-level navigation actions.}

\vspace{-5pt}
\subsection{{Environmental Perception}}
\label{sec:env_percep}
For an agent situated at an arbitrary node in the environment, it receives egocentric views from multiple orientations, positional coordinates, and neighboring nodes.
The LLM Planner focuses on determining the precise location to proceed to in the subsequent step. Hence, there is no need for the LLM to possess information regarding neighboring nodes.

{To facilitate the comprehension of the LLM Planner towards its surroundings, we employ off-the-shelf visual foundation models to obtain scene descriptions, infer the location and identify objects.
Specifically, the field of view (FoV) of each view is set to 90$^{\circ}$, heading angle turns $\theta = 90^{\circ}$ per view from 0$^{\circ}$ to 360$^{\circ}$. The vertical FoV is set to 90$^{\circ}$, resulting in 4 egocentric views for each navigable node.
Subsequently, as depicted in \texttt{Step1} of \Cref{fig:framework}, we employ InternVL~\cite{chen2024internvl} in a two-round conversation. In the first round, the model is informed that the four images are captured from the identical location to facilitate the model to have a holistic comprehension of current environment.
In the second round, the model is tasked with inferring its current position based on the four images and the descriptions generated in the first round.
}
Then following NavGPT~\cite{zhou2023navgpt}, object detection within a range of 3 meters is performed using Faster R-CNN~\cite{faster-rcnn} along with depth provided by the simulator.
The height derived from the simulator is utilized to infer the specific floor on which the agent is located.
{Finally, current observation $\mathcal{O}_{t}$ is formed as:
\textit{Height off ground is \{height\} meters. \{Current position\}. \{Orientation\}: \{Scene description\}. Objects in 3m: \{objects\}.}
}

\vspace{-5pt}
\subsection{{LLM Planning}}
\label{sec:llm_plan}
The LLM Planner is responsible for guiding the Instruction Follower to explore the environment by transforming OOD instructions into fine-grained guidances that are familiar to the Instruction Follower.
{In order to ensure comprehensibility of guidances for the Instruction Follower, we empirically define an action space based on the instructions in the R2R dataset and restrict the LLM Planner to confine the action phrases in guidances within the action space.
Specifically, the action space consists of: \textit{go downstairs/upstairs, go forward, go through, go past, turn around, turn left/right (at), go to, go into, go out of, stop.}}
To obtain the guidance at the $k^{th}$ iteration, comprehensive reasoning is required from the LLM Planner regarding the input instruction $\mathcal{I}$, preceding navigation history $\mathcal{H}$, and current observation $\mathcal{O}_{t}$.
When humans navigate their path, they contemplate three aspects: past trajectory, current position, and future direction.
{To emulate human cognition, the LLM Planner should first review previously explored places to avoid redundant visit, and subsequently employ common sense reasoning to determine the most probable path to the intended destination based on current position and observations from multiple perspectives.
These behaviors are predefined in the system principle, which comprises six parts:}
\begin{enumerate}
\vspace{-2pt}
    \item Role: The LLM is defined as a navigation agent. 
    \item {Objective}: The responsibility of the LLM is to plan the trajectory toward the destination, while another agent (the Instruction Follower, referred to as Agent2) will execute this plan and take it there.
    \item {Input Definitions}: The descriptions of Instruction, Navigation History, and Current Observation.
    \item {Output Requirements}: Requisites of Thought (procedure of decision-making), which Direction to proceed, and Guidance (a fine-grained command where actions should be confined to the given action space).
    \item {Abilities}: The fundamental reasoning structure of the LLM, including summarizing the preceding trajectory, determining next place to explore, and outputting ``Finished!'' to indicate navigation completion, \etc.
    \item {Constraints}: The regulations that the LLM should adhere to, such as ``do not interact with any objects'', \etc.
\end{enumerate}


\vspace{-8pt}
\subsection{{Feasibility Verification}}
\label{sec:feasible}
{
The process of environment perception and LLM planning may encounter errors, such as perceptual hallucinations and the generation of guidance by LLM that instructs the agent to pass through a closed door, which is disallowed in the simulator. 
To ensure that these errors do not affect subsequent navigation, we employ Qwen2-VL~\cite{wang2024qwen2vl} to assess the feasibility of the guidance. 
Specifically, the visual observation is obtained according to the direction output by the LLM Planner. As illustrated in \Cref{fig:framework}, it is fed into the MLLM along with the guidance.
If feasible, proceed to \texttt{Step4} and direct the Instruction Follower to execute the guidance; otherwise, return to \texttt{Step2} and provide feedback on reasons for infeasibility to the LLM Planner for informing new guidance generation.
}

\vspace{-5pt}
\subsection{{Guidance Execution}}
\label{sec:execute}
{
After obtaining the fine-grained guidance, the Instruction Follower is tasked to execute navigational actions to complete the guidance.
The Instruction Follower should possess the capability to strictly adhere to fine-grained instructions.
To enhance accuracy, we propose a multi-model integration mechanism, where three SoTA methods are ensembled as the Instruction Follower: BEVBert~\cite{an2023bevbert} with exceptional local perception, GridMM~\cite{wang2023gridmm} with excellent mapping ability, and ScaleVLN~\cite{wang2023scaling} with superior navigation performance.
As illustrated in \texttt{Step4} in \Cref{fig:framework}, at each time step $t$, if the action decisions of these three models are consistent, they are deemed to accurately follow the guidance, and the Instruction Follower will execute this action. Otherwise, it indicates a complex situation with conflicting model outputs, in which case the Instruction Follower will seek assistance from a LLM.}

{Specifically, the actions are first transformed into textual descriptions, which are presented as choices of a multiple-choice question. Then, GPT-4o-mini is prompted to select an action from the provided options, based on the guidance and navigation history of current iter.
The processes of transforming actions and navigation history into textual descriptions are identical.
}%
Six predefined directional phrases are as follows: \textit{go downstairs, go upstairs, go forward, turn left, turn right, turn around}.
The action is considered as ``go forward'' when the heading rotation is less than 30$^{\circ}$; ``turn left/right'' when the rotation falls between 30$^{\circ}$ and 150$^{\circ}$; otherwise, the action is considered ``turn around''.
When transitioning from $V_t$ to $V_{t+1}$, if the height difference exceeds $0.2m$, then the action is described as ``go downstairs / upstairs''.
In other cases, we randomly select an object that is within a radius of 3 meters of $V_{t+1}$, and describe the action as: 
\textit{\{Directional phrase\} to \{object\}, facing toward \{scene\}.}



\vspace{-5pt}
\subsection{{Object Localization}}
\label{sec:localize}
Besides navigating to a specified location, the agent is also required to locate a target object at the end of the navigation process.
Given that REVERIE provides object annotations at each navigable node, and the bounding boxes in SOON are extracted by an automatic object detector~\cite{anderson2018bottom}, most previous methods extract object bounding boxes at the final node and train a shallow classification network to select an object as the predicted target.
In our work, we add an Object Locator at the end of the navigation process.
{
Specifically, we utilize GPT-4o-mini to extract the target object from the instruction, then employ BLIP-2~\cite{li2023blip2} to compute the similarity between objects at the final node and the target object tokens. The object with the highest similarity is regarded as our designated object.
}

\section{Experimental Setup}
\subsection{Datasets}
The Instruction Follower in our FlexVLN is trained on R2R~\cite{anderson2018vision}. Therefore, to evaluate generalization performance, we focus on three other VLN benchmarks: REVERIE~\cite{qi2020reverie}, SOON~\cite{zhu2021soon} and CVDN-target~\cite{qiao2025copilot}. These benchmarks encompass high-level instructions that exhibit significant differences compared to those in R2R.
\textbf{REVERIE} contains high-level instructions, 
mainly describing the destination and target object. The agent is required to explore the environment, navigate to the goal, and locate the target object without detailed guidance. Instructions contain 21 words on average. The length of expert paths ranges from 4 to 7 steps.
\textbf{SOON} contains more complex instructions describing the target objects, target area, and neighboring areas. The average length of instructions is 47 words. The length of expert paths ranges from 2 to 21 steps with 9.5 steps on average.
{\textbf{CVDN-target} is proposed in VLN-Copilot~\cite{qiao2025copilot}. It only keeps the sentence from the dialog in CVDN~\cite{thomason2020vision} containing the target object as the instruction.}

\vspace{-5pt}
\subsection{Evaluation Metrics}
We utilize the standard evaluation metrics \cite{anderson2018vision,qi2020reverie} for VLN tasks to compare our method with previous approaches.
(1) Trajectory Length (\textbf{TL}): the average path length of all the predicted navigation trajectories in meters.
(2) Navigation Error (\textbf{NE}): the mean distance from the agent's final location to the destination.
(3) Success Rate (\textbf{SR}): the ratio of successful tasks. The task is considered successful if the target object is visible from the agent's final location within a distance of 3 meters.
(4) Oracle SR (\textbf{OSR}): the ratio of tasks of which one of its trajectory nodes can observe the target object within 3 meters.
(5) SR weighted by Path Length (\textbf{SPL}): both the accuracy and efficiency of navigation, which normalizes the success rate with trajectory length. 
(6) Remote Grounding Success (\textbf{RGS}): the ratio of tasks that successfully locate the target object.
(7) RGS weighted by Path Length (\textbf{RGSPL}).
{(8) Goal Progress (\textbf{GP}) measures the difference between completed distance and left distance to the goal.}

\vspace{-5pt}
\subsection{Implementation Details}
{
We apply the GPT API as our LLM Planner, employing ``gpt-4o'' and ``gpt-4o-mini'' for our experimental results.
For environmental perception, we utilize InternVL-40B~\cite{chen2024internvl}} to extract scene descriptions, and Faster R-CNN~\cite{faster-rcnn} to detect objects.
{For feasibility verification, we utilize Qwen2-VL-7B-Instruction~\cite{wang2024qwen2vl}.}
During the collaboration, the LLM Planner can be utilized a maximum of 10 times, while the Instruction Follower is limited to a maximum of 5 movements for each fine-grained guidance generated by the LLM Planner.

\input{experiments}



\section{Conclusion}
In this paper, we propose FlexVLN, which aims to achieve generalization across diverse VLN datasets by introducing a novel hierarchical navigation system. This system seamlessly integrates the reasoning and generalization abilities of the LLM Planner with the fundamental navigation ability of the Instruction Follower.
The LLM Planner transforms OOD instructions into fine-grained guidances that are compatible with the Instruction Follower's capabilities, while the Instruction Follower navigates through the environment based on these guidances instead of OOD instructions.
{Moreover, a verification mechanism and a multi-model integration mechanism are proposed to mitigate infeasible guidances generated by the LLM Planner and improve execution accuracy of the Instruction Follower, respectively.}
Therefore, FlexVLN demonstrates effective generalization capability towards OOD VLN datasets such as REVERIE, SOON, and CVDN-target.
%


\bibliographystyle{IEEEtran}
\bibliography{my}

\begin{IEEEbiography}
[{\includegraphics[width=1in,height=1.25in,clip,keepaspectratio]{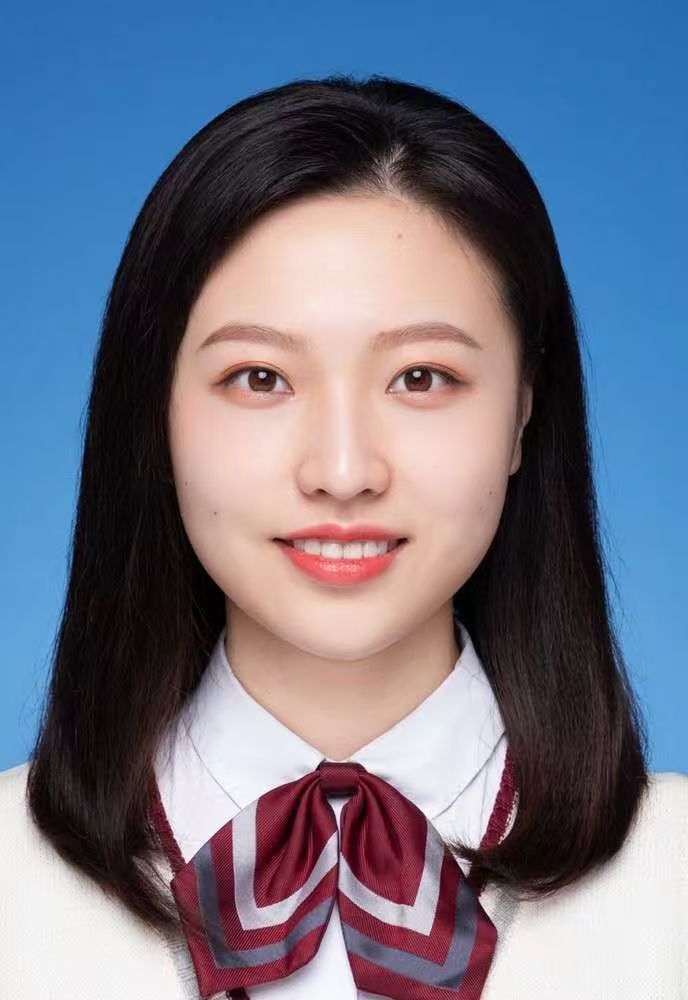}}]{Siqi Zhang}
received her Bachelor's degree in 2018, in Data Science and Big Data Technology from Tongji University, Shanghai, China. She is currently a PhD student at Tongji University. Her research interests include multimodal understanding and embodied intelligence.
\end{IEEEbiography}

\begin{IEEEbiography}
[{\includegraphics[width=1in,height=1.25in,clip,keepaspectratio]{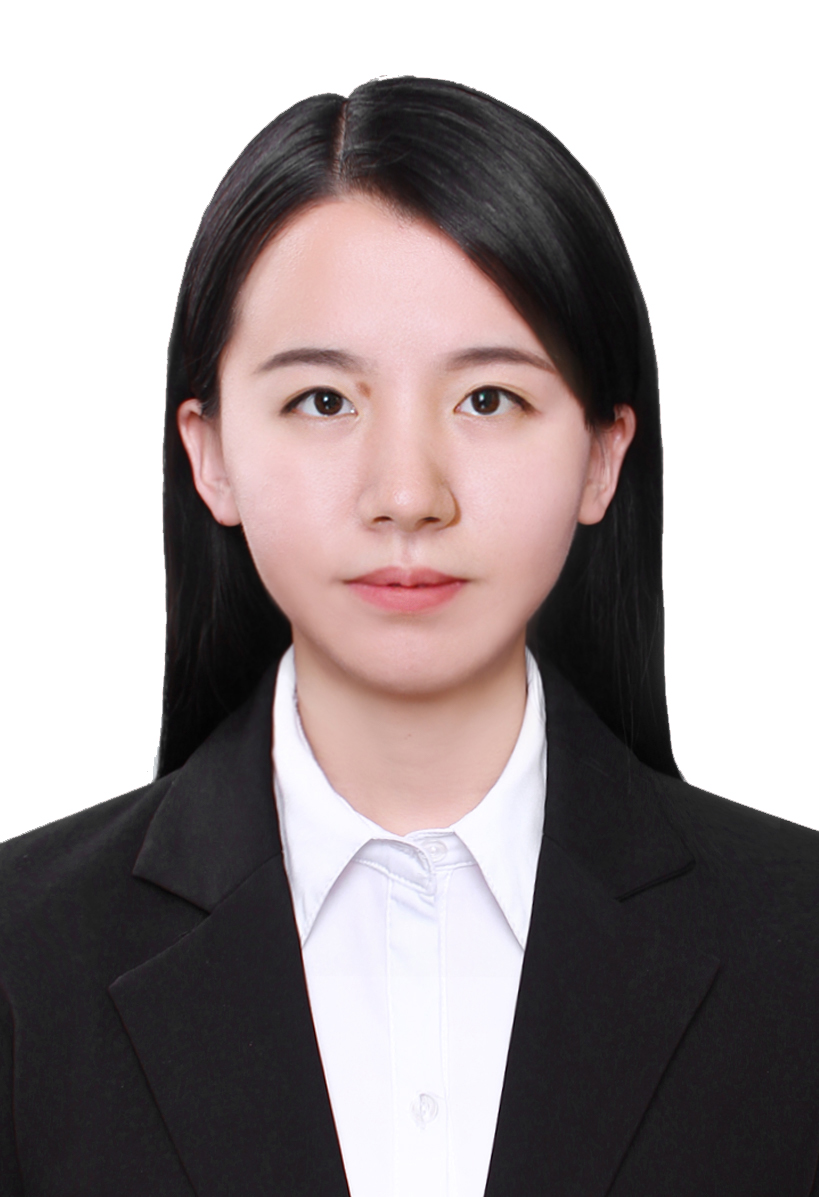}}]{Yanyuan Qiao} receivedd her PhD degree from the University of Adelaide in 2023, her Master’s degree from the University of Chinese Academy of Sciences in 2019, and her Bachelor’s degree from Southeast University in 2016. She is currently a Postdoctoral Research Fellow at the Australian Institute for Machine Learning (AIML), The University of Adelaide. Her research focuses on multimodal learning tasks, including vision-and-language navigation, text-to-image synthesis, and visual question answering. 
\end{IEEEbiography}

\begin{IEEEbiography}
[{\includegraphics[width=1in,height=1.25in,clip,keepaspectratio]{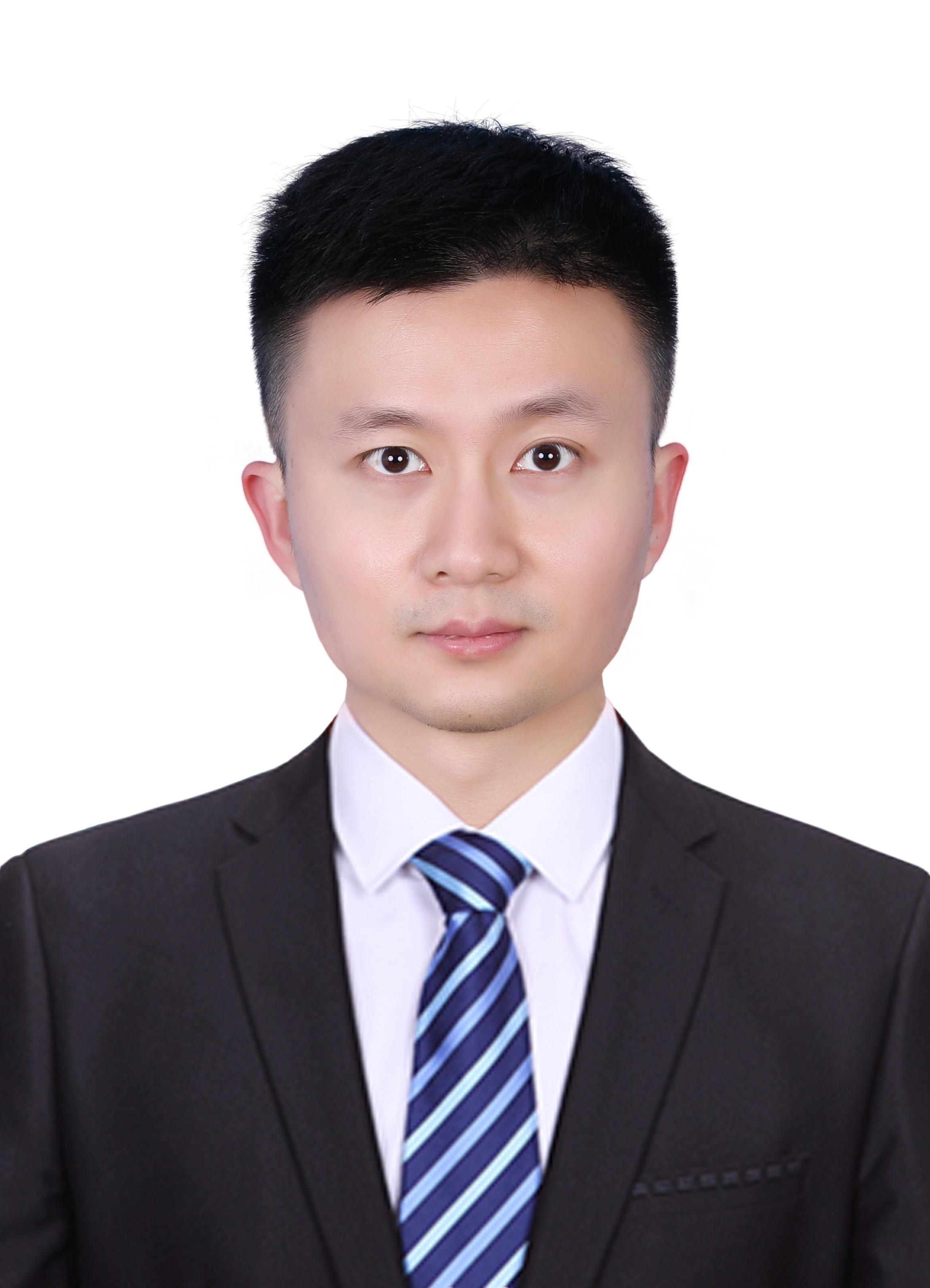}}]{Qunbo Wang}
received his PhD degree in software engineering from Beihang University in 2022. Since 2022, he has been with the Institute of Automation, Chinese Academy of Sciences (CASIA), as an Assistant Professor. His research interests include pre-trained models, weakly supervised learning, embodied intelligence, etc.
\end{IEEEbiography}

\begin{IEEEbiography}
[{\includegraphics[width=1in,height=1.25in,clip,keepaspectratio]{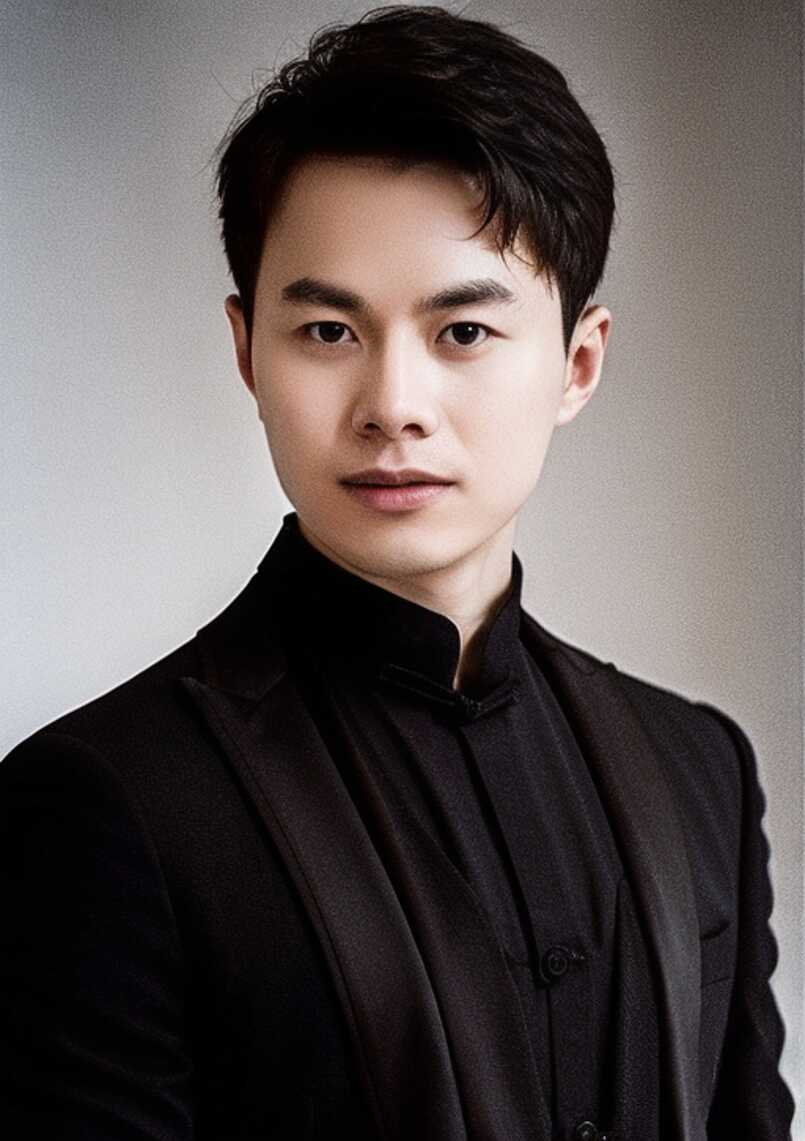}}]{Longteng Guo}
is currently an Associate Professor at the Zidongtaichu Foundation Model Research Center, Institute of Automation, Chinese Academy of Sciences. He obtained his PhD from the same institute in 2021. His research interests encompass deep learning, as well as multimodal understanding and generation.
\end{IEEEbiography}

\begin{IEEEbiography}
[{\includegraphics[width=1in,height=1.25in,clip,keepaspectratio]{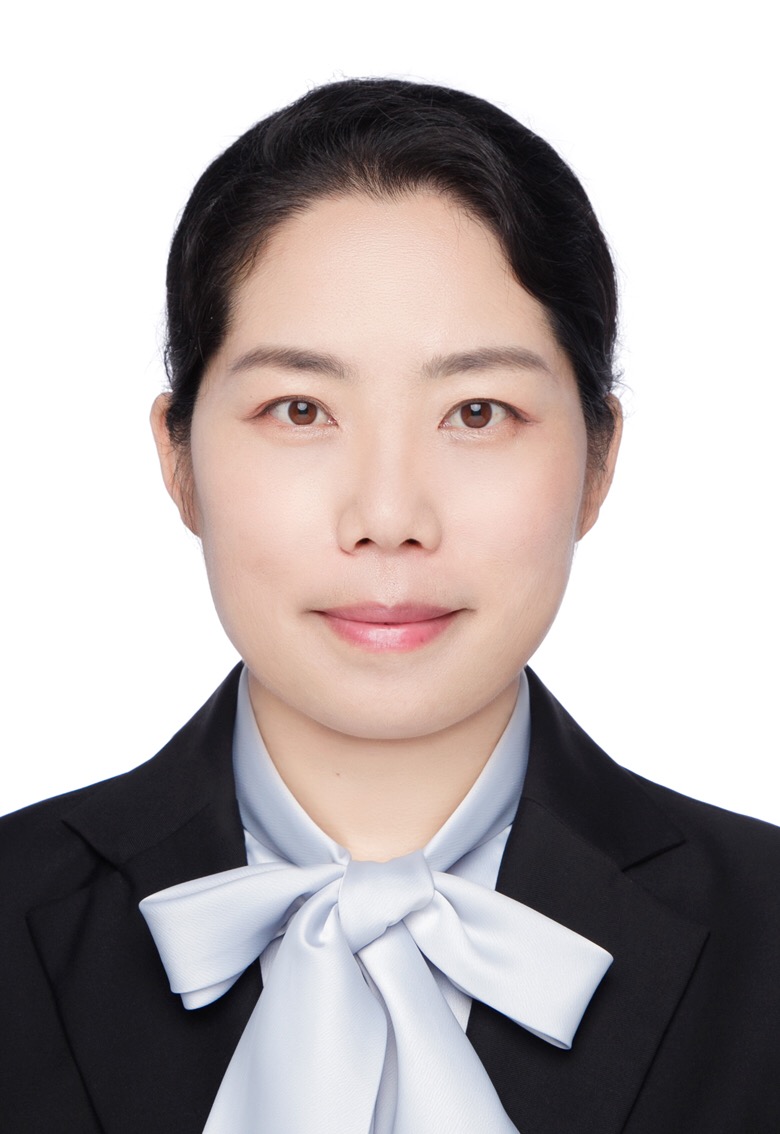}}]{Zhihua Wei}
received dual PhD degree from Tongji University, China and the University of Lyon, France. She is currently a professor and doctoral supervisor at the Development of Computer Science and Technology, Tongji University.
Her research interests encompass natural language processing, multimodal understanding and generation. 
She has authored or co-authored over 100 papers in prestigious international conferences including CVPR, IJCAI, and WWW, as well as leading international journals such as TPAMI. Additionally, she has published two textbooks.
In recent years, she has led seven national projects, including the National Key Research and Development Program and the National Nature Science Foundation.
She has won the Second Prize of the Shanghai Science and Technology Progress Award, and the Wu Wenjun Artificial Intelligence Natural Science Award.
\end{IEEEbiography}

\begin{IEEEbiography}
[{\includegraphics[width=1in,height=1.25in,clip,keepaspectratio]{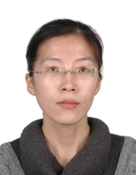}}]{Jing Liu}
received the PhD degree from the Institute of Automation, Chinese Academy of Sciences. Currently, she is a professor at ZIDONGTAICHU Foundation Model Research Center, Institute of Automation, Chinese Academy of Sciences. Her research interests include multimodal foundation models, image content analysis and classification, multimedia information indexing and retrieval, etc.
\end{IEEEbiography}

\end{document}

%% file: experiments.tex
\section{Results and Analysis}

\subsection{Comparison on the REVERIE dataset}
The comparison between our FlexVLN and previous methods on the validation unseen set of the REVERIE dataset is presented in Table~\ref{table:reverie}.
We categorize the previous methods into two types: those specifically trained for REVERIE as in-domain methods, and those not exposed to instructions in REVERIE as out-of-domain methods.
NavGPT~\cite{zhou2023navgpt} can be considered as the LLM-Planner-only baseline.
Since NavGPT is only evaluated on R2R, we replicate its performance using GPT-4o-mini on the validation unseen set of REVERIE.
{BEVBert~\cite{an2023bevbert}, ScaleVLN~\cite{wang2023scaling}, and GridMM~\cite{wang2023gridmm} serve as the Instruction Follower only baseline.
Ensemble refers to the integration of these three models by combining their prediction logits.}
FlexVLN outperforms these baselines to a large extent.

FlexVLN significantly beats all the previous LLM-based methods across all evaluation metrics. These LLM-based methods solely focused on the navigation task while neglecting the crucial task of object location.
The FlexVLN surpasses the latest method MapGPT~\cite{chen2024mapgpt} by a remarkable 11.4\% in terms of SR and an impressive 15.2\% in SPL when MapGPT employs GPT-4, while achieving a notable improvement of 8.2\% in SR and 9.4\% in SPL when MapGPT utilizes GPT-4V.
FlexVLN also outperforms the recent in-domain method LANA~\cite{wang2023lana} by 5.8\% on SR, 7.68\% on RGS and 3.16\% on RGSPL, while achieving comparable SPL.

\begin{table}[t]
\setlength{\tabcolsep}{0.9mm}
\centering
\caption{
{Comparisons on the validation unseen set of REVERIE. $*$: we reproduce NavGPT on REVERIE using GPT-4o-mini. $\dagger$: MapGPT is evaluated on 500 randomly selected samples.}
}
\label{table:reverie}
\vspace{-5pt}
\resizebox{\linewidth}{!}{
\begin{tabular}{ll|ccccc}
\toprule
\multirow{2}{*}{\textbf{Methods}} & \multirow{2}{*}{\textbf{LLM}} & \multicolumn{3}{c}{\textbf{Navigation}} & \multicolumn{2}{c}{\textbf{Grounding}} \\
&  & OSR$\uparrow$ & SR$\uparrow$ & SPL$\uparrow$ & RGS$\uparrow$ & RGSPL$\uparrow$ \\ 
\midrule
\multicolumn{7}{c}{\textbf{\textit{In-Domain}}} \\
\hline

Seq2Seq \cite{anderson2018vision} & -  & 8.07 & 4.20 & 2.84 & 2.16 & 1.63 \\
FAST-MATTN \cite{qi2020reverie} & -  & 28.20 & 14.40 & 7.19 & 7.84 & 4.67 \\
\rvlnbert \cite{hong2020rvlnbert} & -  & 35.02 & 30.67 & 24.90 & 18.77 & 15.27\\
HAMT \cite{chen2021history} & -  & 36.84 & 32.95 & 30.20 & 18.92 & 17.28 \\
DuET \cite{chen2022think} & -  & 51.07 & 46.98 & 33.73 & 32.15 & 23.03 \\
VLN-PETL \cite{qiao2023vln} & - &37.03 & 31.81 & 27.67 & 18.26 & 15.96\\
LANA~\cite{wang2023lana} & -  & 38.54 & 34.00 & 29.26 & 19.03 & 16.18\\
BEVBert~\cite{an2023bevbert} & -  & 56.40 & 51.78 & 36.37 & 34.71 & 24.44 \\
GridMM~\cite{wang2023gridmm} & -  & 57.48 & 51.37 & 36.47 & 34.57 & 24.56 \\
AutoVLN~\cite{chen2022learning} & - & 62.14 & 55.89 & 40.85 & 36.58 & 26.76 \\
ScaleVLN~\cite{wang2023scaling} & -  & 63.85 & 56.97 & 41.84 & 35.76 & 26.05 \\
 
\midrule
\multicolumn{7}{c}{\textbf{\textit{Out-of-Domain}}} \\
\hline
BEVBert (R2R) & - & 52.12 & 26.67 & 20.97 & - & - \\
GridMM (R2R) & - & 52.63 & 30.02 & 24.12 & - & -\\
ScaleVLN (R2R) & -  & 54.50 & 28.60 & 20.61 & - & - \\
Ensemble & - & 53.99 & 31.95 & 26.60 & - & - \\
NavGPT$^*$~\cite{zhou2023navgpt} & GPT-4o-mini  & 24.86 & 9.83 & 6.93 & - & - \\
MapGPT$^\dagger$~\cite{chen2024mapgpt} & GPT-4  & 42.60 & 28.40 & 14.50 & - & - \\
MapGPT$^\dagger$~\cite{chen2024mapgpt} & GPT-4V  & 36.80 & 31.60 & 20.30 & - & - \\

FlexVLN (Ours) & GPT-4o-mini & \textbf{70.59} & 32.94 & 24.34 & 22.75 & 16.10 \\
FlexVLN (Ours) & GPT-4o & {69.53} & \textbf{39.80} & \textbf{29.74} & \textbf{26.71} & \textbf{19.34} \\
\bottomrule
\end{tabular}}
\vspace{-5pt}
\end{table}

\begin{table}[t]
\setlength{\tabcolsep}{1mm}
\centering
\caption{Comparisons on the validation unseen house set of SOON. $*$: we reproduce NavGPT on SOON using GPT-4o-mini.}
\label{table:soon}
\vspace{-5pt}
\begin{tabular}{ll|cccc}
\toprule
\multirow{2}{*}{\textbf{Methods}} & \multirow{2}{*}{\textbf{LLM}} & \multicolumn{3}{c}{\textbf{Navigation}} & \multicolumn{1}{c}{\textbf{Grounding}} \\
&  & OSR$\uparrow$ & SR$\uparrow$ & SPL$\uparrow$ & RGSPL$\uparrow$ \\ 
\midrule
\multicolumn{6}{c}{\textbf{\textit{In-Domain}}} \\
\hline
GBE~\cite{zhu2021soon} & - & 28.54 & 19.52 & 13.34 & 1.16 \\ 
DUET~\cite{chen2022think} & - & 50.91 & 36.28 & 22.58 & 3.75 \\
GridMM~\cite{wang2023gridmm} & - & 53.39 & 37.46 & 24.81 & 3.91 \\
AutoVLN~\cite{chen2022learning} & - & 53.19 & 41.00 & 30.69 & 4.06 \\
\midrule
\multicolumn{6}{c}{\textbf{\textit{Out-of-Domain}}} \\
\hline
BEVBert (R2R) & - & 6.31 & 3.01 & 1.86 & - \\
GridMM (R2R) & - & 7.64 & 4.57 & 2.72 & - \\
ScaleVLN (R2R) & - & 8.11 & 3.57 & 2.24 & - \\
Ensemble & - & 8.67 & 2.86 & 2.25 & - \\
NavGPT & GPT-4o-mini & 12.26 & 2.09 & 1.50 & - \\
FlexVLN & GPT-4o-mini & 29.02 & 13.33 & 9.09 & 4.26 \\
FlexVLN & GPT-4o & \textbf{32.31} & \textbf{23.08} & \textbf{14.96} & \textbf{6.94} \\
\bottomrule
\end{tabular}%
\vspace{-5pt}
\end{table}

\begin{table}[t]
\centering
\caption{Comparison on the validation unseen set of CVDN-target.}
\vspace{-5pt}
\label{table:cvdn}
\begin{tabular}{@{}ll|cc@{}}
\toprule
\textbf{Setting} & \textbf{Methods} & \textbf{TL} & \textbf{GP}$\uparrow$ \\ 
\midrule
\multirow{4}{*}{\textbf{\textit{In-Domain}}} & HAMT & 37.04 & 2.74 \\
 & DUET & 78.53 & 3.70 \\
 & AutoVLN & 61.97 & 3.98 \\
 & VLN-Copilot~\cite{qiao2025copilot} & 40.38 & 4.47 \\ \midrule
\multirow{5}{*}{\textbf{\textit{Out-of-Domain}}} & BEVBert (R2R) & 27.08 & 1.50 \\
 & GridMM (R2R) & 23.80 & 1.34 \\
 & ScaleVLN (R2R) & 28.76 & 2.20 \\
 & Ensemble & 27.83 & 1.90 \\
 & FlexVLN{\tiny GPT-4o-mini} & 29.99 & \textbf{3.63} \\ 
 \bottomrule
\end{tabular}
\vspace{-5pt}
\end{table}

\vspace{-5pt}
\subsection{Comparison on the SOON dataset}
We conduct experiments on SOON validation unseen house set, in which instructions are more complex.
Table~\ref{table:soon} presents the results on the validation unseen house set of SOON.

It should be noted that, given the inherent difficulty of the SOON dataset, our FlexVLN represents the first LLM-based method for SOON and is also a pioneer in testing generalization performance on this benchmark.
FlexVLN achieves superior performance compared to GBE, with an improvement of 3.56\% on SR and 5.78\% on RGSPL.
The performance enhancement of FlexVLN over the Instruction Follower baseline on SOON is more pronounced than that on REVERIE. This suggests that even when a given instruction significantly deviates from the domain of the Instruction Follower, our LLM Planner can still effectively guide the Instruction Follower to complete navigation, thereby demonstrating the generalization capability of our FlexVLN.

\vspace{-5pt}
\subsection{Comparison on the CVDN-target dataset}
{
\Cref{table:cvdn} presents results on the validation unseen set of CVDN-target, where only the target object is included in the instruction, thereby increasing the difficulty of the task.
Considering the API costs, we only report the result of FlexVLN with GPT-4o-mini.
It can be observed that FlexVLN achieves superior performance compared to HAMT, with an increase of 0.89 in GP, while achieving comparable performance with DUET.
Moreover, FlexVLN demonstrates notable performance enhancements across all the Instruction Follower baselines, showcasing its exceptional generalization performance.
}

\begin{table}[t]
\centering
\caption{Ablation on feasibility verification and confining action phrases.}
\vspace{-5pt}
\label{table:ab1}
\begin{tabular}{ll|cccc}
\toprule
& Methods  & NE$\downarrow$ & OSR$\uparrow$ & SR$\uparrow$ & SPL$\uparrow$ \\
\midrule
{\romannumeral1}& NavGPT  & 8.29 & 34 & 11 & 7.46 \\
{\romannumeral2}& FlexVLN  & 6.91 & 76 & 37 & 26.38\\
\midrule
{\romannumeral3}& w/o verification & 7.40 & 74 & 34 & 22.83 \\
{\romannumeral4}& w/o action space & 7.46 & 75 & 30 & 20.88 \\
\bottomrule
\end{tabular}
\vspace{-5pt}
\end{table}

\begin{table}[t]
\centering
\caption{Comparison between the original R2R instructions and LLM rewrite instructions.}
\vspace{-5pt}
\begin{tabular}{@{}l|ccc|ccc@{}}
\toprule
\multirow{2}{*}{Model} & \multicolumn{3}{c|}{R2R instructions} & \multicolumn{3}{c}{Rewritten instructions} \\
 & OSR$\uparrow$ & SR$\uparrow$ & SPL$\uparrow$ & OSR$\uparrow$ & SR$\uparrow$ & SPL$\uparrow$ \\
\midrule
BEVBert~\cite{an2023bevbert} & 84 & 75 & 64 & 83 & 74 & 62 \\
GridMM~\cite{wang2023gridmm} & - & 75 & 64 & 82 & 73 & 62 \\
ScaleVLN~\cite{wang2023scaling} & 88 & 81 & 70 & 88 & 80 & 68 \\
\bottomrule
\end{tabular}
\label{table:rewrite}
\vspace{-5pt}
\end{table}

\begin{table}[t]
\centering
\caption{Comparison of various ways of environmental perception.}
\label{table:visual}
\vspace{-5pt}
\begin{tabular}{@{}ll|cccc@{}}
\toprule
& Perception Method & NE$\downarrow$ & OSR$\uparrow$ & SR$\uparrow$ & SPL$\uparrow$ \\
\midrule
{\romannumeral1} & 8 views, BLIP-2 & 8.20 & 71 & 28 & 19.82 \\
{\romannumeral2} & 4 views, InternVL (Ours) & 6.91 & 76 & 37 & 26.38 \\
\midrule
{\romannumeral3} & w/o position & 7.76 & 72 & 34 & 24.77 \\
\bottomrule
\end{tabular}%
\vspace{-10pt}
\end{table}

\begin{table}[t]
\centering
\caption{Comparison of various ways of feeding guidances to the Instruction Follower. ``Map'' refers to the topological memory map built during the previous guidances execution. ``single'' denotes only the current guidance is provided, while ``multi'' indicates that all guidances up to the current step are concatenated and provided.}
\label{table:Instruction Follower-v1}
\vspace{-5pt}
\begin{tabular}{@{}cc|cccc@{}}
\toprule
Map & Guidance & NE$\downarrow$ & OSR$\uparrow$ & SR$\uparrow$ & SPL$\uparrow$ \\ \midrule
\multirow{2}{*}{\ding{56}} & single & 7.54 & 65 & 20 & 13.97 \\
 & multi & \textbf{6.54} & 72 & 29 & 17.16 \\ 
\midrule
\multirow{2}{*}{\ding{52}} & single & 7.07 & 75 & 30 & 21.97 \\
 & multi & 6.91 & \textbf{76} & \textbf{37} & \textbf{26.38} \\ 
\bottomrule
\end{tabular}%
\vspace{-5pt}
\end{table}

\begin{table}[t]
\centering
\caption{Ablation of the navigation history. Guidance refers to the previously generated guidances. Trajectory denotes the textual descriptions of the Instruction Follower's trajectory. $\#1$ and $\#2$ are two types of trajectory descriptions.}
\label{table:nav_his}
\vspace{-5pt}
\begin{tabular}{@{}lcc|cccc@{}}
\toprule
& Guidance & Trajectory & NE$\downarrow$ & OSR$\uparrow$ & SR$\uparrow$ & SPL$\uparrow$ \\ 
\midrule
{\romannumeral1}& - & - & 7.80 & 73 & 28 & 21.42 \\
{\romannumeral2}& \ding{52} & $\#1$ & 6.72 & 72 & 33 & 23.49 \\
{\romannumeral3}& - & $\#1$ & 7.11 & 75 & 33 & 24.16 \\
{\romannumeral4}& \ding{52} & $\#2$ & 6.70 & 74 & 34 & 25.05 \\
{\romannumeral5}& - & $\#2$ & 6.91 & 76 & 37 & 26.38 \\ 
 \bottomrule
\end{tabular}%
\vspace{-5pt}
\end{table}

\begin{table}[t]
\setlength{\tabcolsep}{1mm}
\centering
\caption{Comparison of various Instruction Followers. All the results are evaluated on the validation unseen set of REVERIE.}
\vspace{-5pt}
\begin{tabular}{@{}l|ccc|cccccc@{}}
\toprule
\multirow{3}{*}{Model} & \multicolumn{3}{c|}{\multirow{2}{*}{Train on REVERIE}} & \multicolumn{6}{c}{Train on R2R} \\ \cmidrule(l){5-10} 
 & \multicolumn{3}{c|}{} & \multicolumn{3}{c|}{No finetuning} & \multicolumn{3}{c}{with FlexVLN} \\ \cmidrule(l){2-10} 
 & OSR$\uparrow$ & SR$\uparrow$ & SPL$\uparrow$ & OSR$\uparrow$ & SR$\uparrow$ & \multicolumn{1}{c|}{SPL$\uparrow$} & OSR$\uparrow$ & SR$\uparrow$ & SPL$\uparrow$ \\ \midrule

AZHP & 53.65 & 48.31 & 36.63 & 62 & 29 & \multicolumn{1}{c|}{24.73} & 64 & 31 & 26.08 \\
BEVBert & 56.40 & 51.78 & 36.37 & 54 & 24 & \multicolumn{1}{c|}{18.18} & 67 & 33 & 25.62 \\
GridMM & 57.48 & 51.37 & 36.47 & 54 & 30 & \multicolumn{1}{c|}{22.67} & 62 & 32 & 26.63 \\
ScaleVLN & 63.85 & 56.97 & 41.84 & 55 & 26 & \multicolumn{1}{c|}{19.42} & 65 & 31 & 24.39 \\ 
Ensemble & - & - & - & 52 & 27 & \multicolumn{1}{c|}{22.23} & 64 & 34 & 28.21 \\
\bottomrule
\end{tabular}
\label{tab:IF}
\vspace{-5pt}
\end{table}

\begin{table}[t]
    \centering
    \caption{Comparison of various (M)LLM Planners.}
    \vspace{-5pt}
    \begin{tabular}{ll|cccc}
    \toprule
        \multicolumn{2}{c|}{Planner} & NE$\downarrow$ & OSR$\uparrow$ & SR$\uparrow$ & SPL$\uparrow$ \\
    \midrule
        \multirow{2}{*}{LLM} & GPT-4o-mini & 6.91 & 76 & 37 & 26.38 \\
        & GPT-4o & 6.71 & 67 & 39 & 29.82 \\
    \midrule
        MLLM & GPT-4o-mini & 7.15 & 73 & 34 & 24.87 \\
    \bottomrule
    \end{tabular}
    \label{table:llm-p}
\vspace{-5pt}
\end{table}

\begin{table}[t]
    \centering
    \caption{Comparison of various object locators on the validation unseen set of REVERIE.}
    \vspace{-5pt}
    \label{table:locator}
    \begin{tabular}{l|ccc}
    \toprule
        Model & SR$\uparrow$ & RGS$\uparrow$ & RGSPL$\uparrow$ \\
    \midrule
        CLIP~\cite{radford2021learning} & 39.80 & 25.51 & 18.43 \\
        BLIP-2~\cite{li2023blip2} & 39.80 & 26.71 & 19.43 \\
    \bottomrule
    \end{tabular}
    \label{tab:my_label}
\vspace{-5pt}
\end{table}

\vspace{-5pt}
\subsection{Ablation Studies}
In ablation experiments, we construct a baseline with {GPT-4o-mini} for its easier access and budget-friendly costs.
Regarding building validation scenarios, NavGPT~\cite{zhou2023navgpt} constructs the ablation set by sampling one trajectory from each of the 72 environments.
However, the Instruction Follower has been exposed to the 60 environments in the training set.
Therefore, we construct a new validation split by sampling from the original unseen validation set of REVERIE, which consists of 10 scenes. From each scene, one trajectory and its corresponding instruction are randomly selected.
In total, we sample 100 instances to conduct the ablation study.
{To ensure the reliability of the ablation set, the performance of NavGPT when using GPT-4o-mini is included in \Cref{table:ab1}.}

\vspace{2pt}
\noindent\textbf{1) The effect of feasibility verification.}
{The necessity of incorporating a MLLM for verifying the feasibility of guidance generated by the LLM Planner is demonstrated in \Cref{table:ab1}.
The comparison between Line {\romannumeral2} and Line {\romannumeral3} reveals that the absence of feasibility verification would result in a decrease of 3\% in SR and 3.55\% in SPL.
This indicates that feasibility verification in \texttt{Step 3} can effectively mitigate the negative impact caused by the errors occurring during the environmental perception and LLM planning processes.
}

\vspace{2pt}
\noindent\textbf{2) The effect of confining action phrases.}
{In order to enhance the comprehensibility of the guidance generated by the LLM Planner for the Instruction Follower, an action space is predefined and the action phrases in the guidance are constrained within this action space.
To ensure that the Instruction Follower can comprehend the actions within the action space, we rewrite the original instructions in the R2R dataset with GPT-4o-mini, constraining the action phrases within this space. \Cref{table:rewrite} represents the comparison of three models in the Instruction Follower on both the original R2R instructions and their rewritten counterparts.
It can be observed that their performances are comparable. Slight degradation may occur due to errors introduced by LLM during rewriting, such as altering the order of actions in the instruction.
In \Cref{table:rewrite}, the comparison between Line {\romannumeral2} and Line {\romannumeral4} demonstrates the necessity of constraining action phrases when generating guidances.
}

\vspace{2pt}
\noindent\textbf{3) Environmental perception.}
{The effects of different environmental perception methods on navigation performance are compared in \Cref{table:visual}.
We apply the perception method in NavGPT in Line {\romannumeral1}, where BLIP-2~\cite{li2023blip2} is utilized to extract image descriptions from 24 views under 3 distinct elevations. A LLM is then applied to summarize the descriptions of various elevations within the same heading, resulting in 8 view descriptions. However, due to the separate extraction of each view's description, a comprehensive understanding of the scene is lacking. For instance, in a room with tables and chairs, some views are described as office paper while others are labeled as living room, making it challenging for the LLM Planner to determine the current location accurately.
Comparatively, FlexVLN enables the InternVL to generate descriptions for each view based on panoramic observations and then infer the current position. Comparison between Line {\romannumeral1} and Line {\romannumeral2} demonstrates the effectiveness of our method, leading to an increase in SR by 9\% and SPL by 6.56\% respectively.
Moreover, in Line {\romannumeral3}, when excluding the prediction of current position, there is a decrease in SR by 2\% and SPL by 1.61\%, validating that the prediction of current location contributes to enhancing the LLM Planner's understanding of the environment and facilitating path planning.
}

\vspace{2pt}
\noindent\textbf{4) How to guide the Instruction Follower?}
To investigate the optimal format for providing fine-grained guidance to the Instruction Follower in order to maximize its potential and thereby enhance the overall success rate of FlexVLN, we conduct an ablation study. This study aims to explore whether the Instruction Follower should receive each guidance generated by the LLM Planner individually or concatenate it with previous guidances, as well as whether to retain the topological memory map constructed during the execution of previous guidances.
The results presented in \Cref{table:Instruction Follower-v1} demonstrate the crucial role of topological memory mapping in instruction following. Moreover, the Instruction Follower can align the trajectory and instruction, thus incorporating previous guidances is more reasonable for it. 

\vspace{2pt}
\noindent\textbf{5) Description of navigation history.}
The navigation history aims to provide the LLM Planner with information regarding previously explored locations, enabling it to comprehend the spatial layout of these places, and thereby facilitating more effective planning for future actions.
We ablate on ``Guidance'' (previous guidances from the LLM Planner) and ``Trajectory'' (moving trajectory of the Instruction Follower).
The efficacy of providing the LLM Planner with navigation history can be demonstrated by the results in Line {\romannumeral1}.
In terms of depicting the trajectory of the Instruction Follower, we consider two types.
Trajectory $\#1$ is a symbolic representation denoting \textit{``Turn \{turning angle\} degrees. Move \{distance\} meters.''}
Trajectory $\#2$ is the one applied in FlexVLN, namely \textit{``\{directional pharse\} to \{object\}, facing toward \{scene\}''}.
As demonstrated in \Cref{table:nav_his}, trajectory $\#2$ exhibits superior performance compared to $\#1$ (Line {\romannumeral2} versus Line {\romannumeral4}, and Line {\romannumeral3} versus Line {\romannumeral5}). 
This discrepancy may arise from the limited capability of the LLM Planner to comprehend the symbolic descriptions, while trajectory $\#2$ incorporates landmarks and scene descriptions, potentially enhancing the LLM Planner's understanding of previously explored areas.
{The combination of previous guidances, however, impairs the navigation performance (Line {\romannumeral2} versus Line {\romannumeral3}, and Line {\romannumeral4} versus Line {\romannumeral5}).
This may result in the inevitable inconsistency between guidance and the actual execution of the Instruction Follower. The discrepancy may affect the LLM Planner's comprehension of navigation history upon joining previous guidances.}

\vspace{2pt}
\noindent\textbf{6) Instruction Follower.}
{\Cref{tab:IF} presents the comparison of various Instruction Followers, showcasing the performance of in-domain training on REVERIE, directly transferring to REVERIE after training on R2R, and transferring to REVERIE with FlexVLN after training on R2R.
Notably, all models exhibit enhanced performance with FlexVLN, especially BEVBert which shows a 7\% increase in SR and a 7.44\% increase in SPL. The improvement can likely be attributed to its exceptional local perception capability, enabling it to execute fine-grained instructions more accurately.
``Ensemble, with FlexVLN'' refers to combining prediction logits from the three models for action decision-making during guidance execution in \texttt{Step 4} without asking LLM for help. This result also demonstrates that seeking LLM assistance when decisions are inconsistent aids accurate guidance execution of the Instruction Follower.
}

\vspace{2pt}
\noindent\textbf{7) LLM Planner.}
As FlexVLN relies on the generalization and reasoning capabilities of the LLM Planner, a comparative analysis of various LLMs is conducted in \Cref{table:llm-p}. Moreover, considering the inevitable loss of information during the conversion from visual observations to textual descriptions, we also employ MLLM as the planner. 
{It can be observed that utilizing GPT-4o-mini as MLLM Planner results in a degradation in performance. This could potentially be attributed to the fact that MLLM Planner is required to simultaneously accomplish environmental perception, current position inference, and path planning.
In comparison to LLM Planner, MLLM Planner is burdened with more tasks, thereby resulting in a decline in performance.
Therefore, GPT-4o is adopted as the preferred LLM Planner of our FlexVLN.}

\vspace{2pt}
\noindent\textbf{8) Object Locator.}
{
The comparison of two object locators is presented in \Cref{table:locator}. It can be observed that BLIP-2 outperforms CLIP in locating objects, exhibiting a 1.2\% higher RGS and 1\% higher RGSPL than CLIP on the REVERIE validation unseen set.
}

\begin{table}[t]
\centering
\caption{Comparison of LLM calls when using GPT-4o-mini on the ablation set.}
\vspace{-5pt}
\label{table:cost}
\begin{tabular}{l|cccc}
\toprule
Methods & LLM calls & Total Tokens & Cost (\$) & Inf. Time (min) \\
\midrule
NavGPT & 1,430 & 3,288,733 & 0.58 & 70 \\
FlexVLN & 692 & 405,853 & 0.08 & 34\\
\bottomrule
\end{tabular}
\vspace{-5pt}
\end{table}

\begin{figure*}
    \centering
    \includegraphics[width=\linewidth]{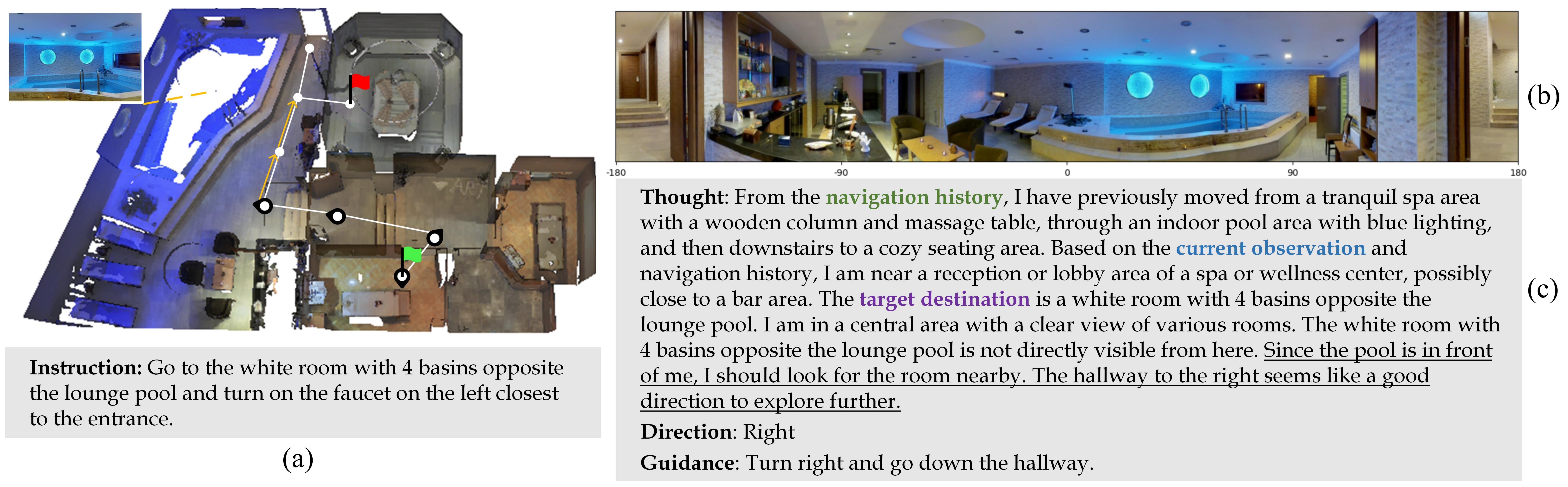}
    \vspace{-20pt}
    \caption{An example on the REVERIE dataset illustrates the collaboration between the LLM Planner and Instruction Follower.}
    \label{fig:visualization}
\end{figure*}

\begin{figure*}
    \centering
    \includegraphics[width=\linewidth]{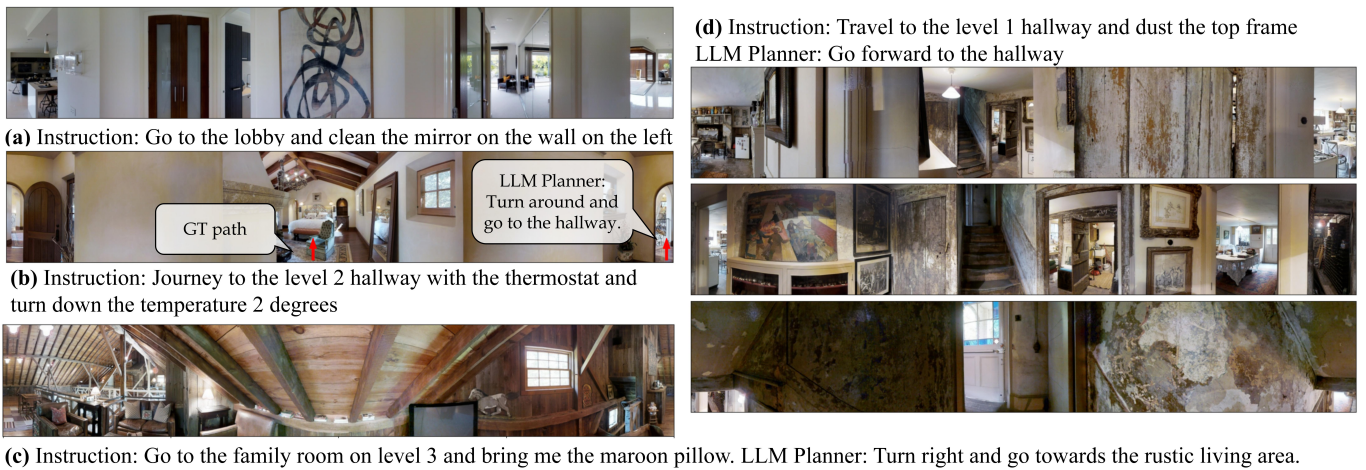}
    \vspace{-20pt}
    \caption{{Visualization of error cases.}}
    \label{fig:error}
\vspace{-12pt}
\end{figure*}

\vspace{-5pt}
\subsection{Analysis}
\noindent\textbf{Cost analysis.}
{
\Cref{table:cost} presents the comparison of the cost required to complete the inference on the ablation set between NavGPT and FlexVLN when both using GPT-4o-mini. All the metrics are calculated with LangSmith\footnote{https://www.langchain.com/langsmith} toolkit.
It can be observed that FlexVLN only invokes LLM for 48\% times of NavGPT, resulting in a reduction of cost to merely 14\% of NavGPT and halving the inference time. This substantiates our FlexVLN's capability not only to significantly enhance generalization but also to substantially diminish the frequency of LLM calls.
}

\vspace{2pt}
\noindent\textbf{Case study.}
In Figure~\ref{fig:visualization}, we showcase an example from REVERIE to demonstrate the integration of high-level planning of the LLM Planner and low-level navigation action execution of the Instruction Follower.
During the reasoning process of the LLM Planner, it first summarizes the navigation history (green words), determines the current position (blue words), analyzes the target location (purple words).
Subsequently, it employs commonsense reasoning to ascertain the next place to explore along with its underlying rationale (underlined words).
Following the guidance generated by the LLM Planner, as depicted by the yellow path in Figure~\ref{fig:visualization}(a), the Instruction Follower executes a sequence of four navigation actions to reach the swimming pool area, which is close to the intended destination.

\vspace{2pt}
\noindent\textbf{Failure analysis.}
{
We randomly sample 100 error samples from the results on the validation unseen set of REVERIE for analysis of error cases and presented some cases in \Cref{fig:error}.
Among them, 37 samples are identified as challenging instances. For example, in (a), even human beings find it difficult to determine the appropriate path. In (b), although the guidance generated by the LLM Planner appears reasonable, there is a deviation from the ground-truth trajectory.
Additionally, 23 samples are attributed to errors made by the LLM Planner. In (c), despite the maroon pillow being visible on the left side, the LLM Planner incorrectly suggests moving towards the right. This is due to insufficient environmental perception and inadequate description of objects.
The remaining 40 samples are determined to be caused by errors made by the Instruction Follower. In (d), although LLM Planner generated guidance instructing ``go forward to the hallway'', the Instruction Follower failed to stop after entering the hallway and proceeded upstairs instead, resulting in the execution error.
}